\documentclass[conference]{IEEEtran}
\IEEEoverridecommandlockouts
\usepackage{cite}
\usepackage{amsmath,amssymb,amsfonts}
\usepackage{algorithmic}
\usepackage{graphicx}
\usepackage{textcomp}
\usepackage{xcolor}
\def\BibTeX{{\rm B\kern-.05em{\sc i\kern-.025em b}\kern-.08em
    T\kern-.1667em\lower.7ex\hbox{E}\kern-.125emX}}

\usepackage{latexsym}
\usepackage{amssymb}
\usepackage{amsmath}
\usepackage{amsthm}
\usepackage{booktabs}
\usepackage{enumitem}
\usepackage{graphicx}
\usepackage{color}

\usepackage{algorithm}
\usepackage{algorithmic}
\usepackage{amsmath}
\usepackage{multirow}
\usepackage{colortbl}
\usepackage{booktabs} 
\usepackage{bm}
\usepackage{amssymb}
\newcolumntype{M}[1]{>{\centering\arraybackslash}p{#1}}

\newcommand{\eg}{e.g.}
\newcommand{\ie}{i.e.}

\begin{document}

\title{Graph Neural Networks as a Substitute for Transformers in Single-Cell Transcriptomics\\
\thanks{This work was initially completed in June 2024.}
}

\author{
\IEEEauthorblockN{Jiaxin Qi\IEEEauthorrefmark{1},
Yan Cui\IEEEauthorrefmark{2},
Jinli Ou\IEEEauthorrefmark{2},
Jianqiang Huang\IEEEauthorrefmark{1},
Gaogang Xie\IEEEauthorrefmark{1}}
\IEEEauthorblockA{\IEEEauthorrefmark{1}Computer Network Information Center, Chinese Academy of Sciences, Beijing, China \\
Email: \{jxqi, jqhuang, xie\}@cnic.cn}
\IEEEauthorblockA{\IEEEauthorrefmark{2}Hangzhou Institute for Advanced Study, University of Chinese Academy of Sciences, Hangzhou, China \\
Email: cuiyan.ch@gmail.com, oujinli@zuaa.zju.edu.cn}
}


\maketitle

\begin{abstract}
Graph Neural Networks (GNNs) and Transformers share significant similarities in their encoding strategies for interacting with features from nodes of interest, where Transformers use query-key scores and GNNs use edges. Compared to GNNs, which are unable to encode relative positions, Transformers leverage dynamic attention capabilities to better represent relative relationships, thereby becoming the standard backbones in large-scale sequential pre-training. However, the subtle difference prompts us to consider: if positions are no longer crucial, could we substitute Transformers with Graph Neural Networks in some fields such as Single-Cell Transcriptomics? In this paper, we first explore the similarities and differences between GNNs and Transformers, specifically in terms of relative positions. Additionally, we design a synthetic example to illustrate their equivalence where there are no relative positions between tokens in the sample. Finally, we conduct extensive experiments on a large-scale position-agnostic dataset—single-cell transcriptomics—finding that GNNs achieve competitive performance compared to Transformers while consuming fewer computation resources. These findings provide novel insights for researchers in the field of single-cell transcriptomics, challenging the prevailing notion that the Transformer is always the optimum choice.

\end{abstract}  


\section{Introduction}

\begin{figure*}[ht]
\centering\includegraphics[width=\linewidth]{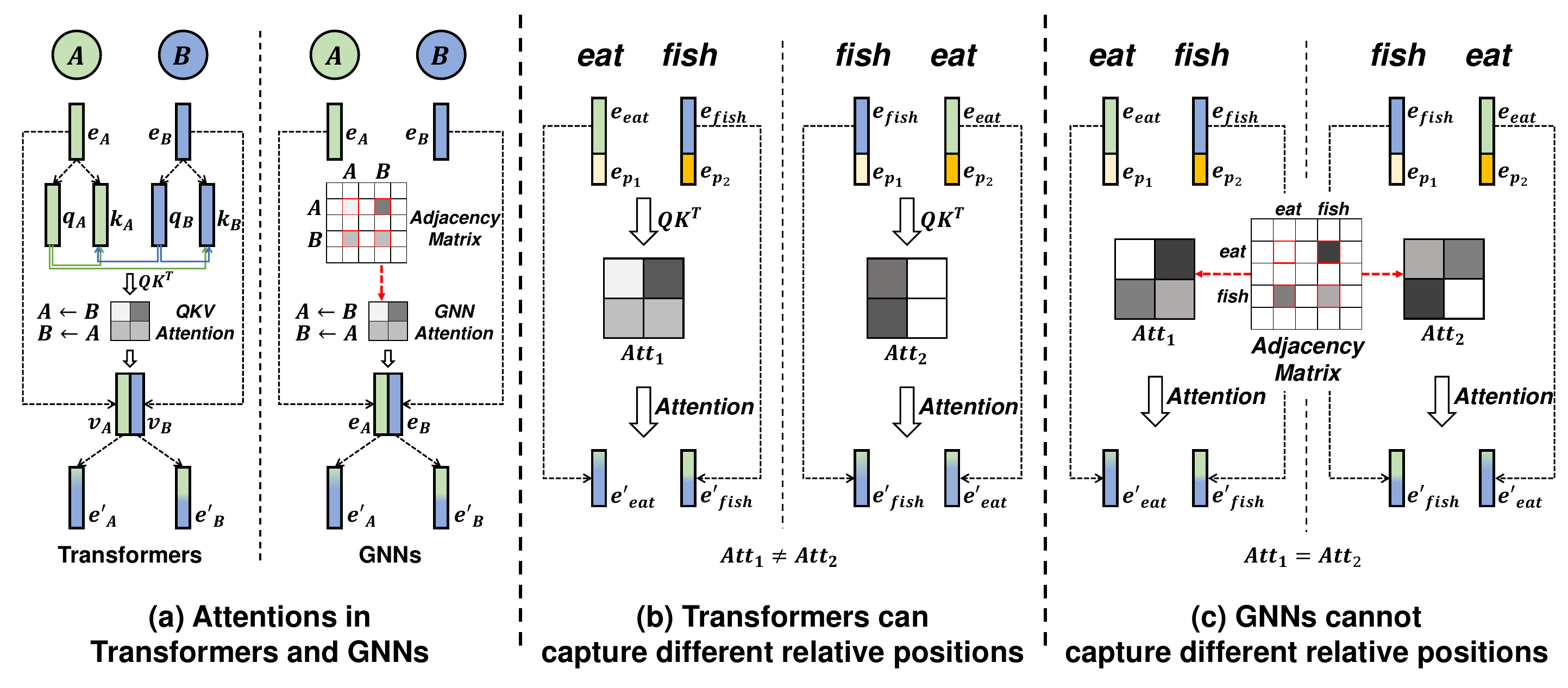}
    \caption{Comparisons of Transformers and GNNs. (a) Similarities: Both of them extract features through node interactions, where Transformers use QKV attention and GNNs use adjacency matrix, which can be considered as attention. (b) and (c) illustrate the differences: When dealing with varying relative positions between nodes (\eg, \texttt{eat} and \texttt{fish}), Transformers generate distinct attention weights, \ie, $Att_1\neq Att_2$, to describe the variations, while GNNs fail to capture different relative positions due to the static attention weights retrieved from adjacency matrix, \ie, $Att_1=Att_2$ (after symmetry).}
    \label{fig:teaser}
\end{figure*}

Recent developments have increasingly highlighted the similarities between Transformers and Graph Neural Networks (GNNs)~\cite{velivckovic2023everything}, particularly in their mechanisms for refining features across nodes (\ie, tokens). 
Transformers~\cite{vaswani2017attention}, the most popular model for handling sequences of nodes, such as natural language and single-cell transcriptomics, employ an attention strategy termed Query-Key-Value (QKV) mechanism to dynamically transfer features across nodes. Specifically, for each node, Transformers compute the soft attention scores by multiplying its query with the keys of other nodes, thereby using attention to weightedly accumulate the values from others. 
Similarly, vanilla GNNs~\cite{scarselli2008graph}, another popular architecture for processing nodes, utilize adjacency matrices to propagate information between nodes. Specifically, for each node, GNNs calculate a weighted sum of features from neighboring nodes, as dictated by the edge weights associated with the node. As shown in Figure~\ref{fig:teaser}(a), the adjacency matrix in GNNs can be considered as a global and static attention map, which is functionally similar to the QKV attention in Transformers. 
This illustrates the fundamental similarities in the feature extraction principles in GNNs and Transformers.

However, in the domain of sequential processing, particularly in large-scale language modeling, Transformers are consistently recognized as the standard architecture~\cite{brown2020languageGPT3,touvron2023llama,zhang2022opt}, while GNNs are seldom considered.
The primary reason is that GNNs are unable to encode \textit{relative positions} (Note that, the proposed relative positions is not the position encoding which could also be achieved by GNNs), which is critical for processing sequential data.
Unlike absolute positions, which can be straightforwardly encoded into positional embeddings, relative positions describe the ordering of nodes, which fundamentally changes the information transfer mechanism between nodes, which is much more challenging to capture. 
For example, as illustrated in Figure~\ref{fig:teaser}(b), when processing the sequences such as \texttt{eat} \texttt{fish} versus \texttt{fish} \texttt{eat}, the shift of relative positions turns a verb-object to a subject-verb relationship between the two nodes. Consequently, for the node \texttt{eat}, the interested features received from the node \texttt{fish} will change in response to the altered relative relationships, which can be implemented as varying attention weights. 
Thanks to the QKV mechanism, Transformers compute attention weights based on the current node features, and thus the changes of relative positions for \texttt{eat} and \texttt{fish} can be captured by the different attention calculations. 
In contrast, GNNs rely on the static adjacency matrix to assign uniform attention weights between nodes, failing to adapt when relative positions shift.
As shown in Figure~\ref{fig:teaser}(c), the interaction between \texttt{eat} and \texttt{fish} is mediated by attention weights retrieved from the adjacency matrix, which is used to process both \texttt{eat} \texttt{fish} and \texttt{fish} \texttt{eat}. Therefore, the inherent limitations of GNNs in adapting to varied relative positions render them naturally inferior in sequential encoding.

Given the aforementioned distinctions, we are prompted to explore whether GNNs are equivalent to Transformers when positions are absent, such as under the single-cell transcriptomics settings, \ie when encoding for varied relative positions is unnecessary. Considering that compared to resource-intensive Transformers~\cite{heaton2018ian,hu2021lora}, GNNs do not require real-time calculation for dynamic attentions, which allows for reductions in computation burden and complexity, \ie, reducing GPU computation and memory requirements. Therefore, using GNNs as substitutes for Transformers when they are equivalent presents a promising approach for the datasets where there is no relative positions, \eg, single-cell transcriptomics.

In this work, we substantiate the equivalence between GNNs and Transformers without positions through a comprehensive discussion that encompasses theoretical analysis, synthetic experiments, and extensive large-scale training. First, we clarify the settings of sequence-like data without positions and formulate how GNNs and Transformers handle interactions across nodes, thereby formally demonstrating their conceptual unity. Based on preliminary theoretical deductions, we establish the foundation for the equivalence between the two architectures. Second, we design a synthetic experiment where the relative positions can be explicitly encoded into the dataset generation, allowing us to compare the optimization difference between GNNs and Transformers when trained on the datasets with or without relative positions. As illustrated in Table~\ref{toy_table} and Figure~\ref{fig:loss_toy}, we find that the superiority of Transformers is from the relative position encoding and reveal that their optimizations are equivalent in the absence of such effects. Finally, to further validate the equivalence, we conduct extensive experiments on a large-scale position-agnostic dataset, single-cell transcriptomics. 
The experimental results demonstrate that GNNs could achieve competitive performance compared with Transformers under the same parameter settings, while only requiring about $1/8$ of the memory and about $1/4$ to $1/2$ of the computational resources in our implementation. In single-cell transcriptomics, characterized by their considerable sample sizes and large sequence length, traditional methods~\cite{scGPT,scFoundation} employing Transformers consistently suffer significant computational demands. Our findings introduce a novel perspective to the community, suggesting that GNNs may serve as an alternative architecture to Transformers for single-cell transcriptomics.

Our main contributions can be summarized as follows:
\begin{enumerate}[topsep=1.5pt, partopsep=0pt, itemsep=0.5ex, parsep=0pt]
    \item We formalize GNNs and Transformers to show the fundamental consistency and difference between them and demonstrate their equivalence without positions.
    \item We design a synthetic example to explicitly show the effects of relative positions, which confirms that the superiority of Transformers stems from relative position encoding, and also shows that GNNs and Transformers are equivalent when relative positions are removed.
    \item We conduct extensive experiments on a large-scale position-agnostic dataset, single-cell transcriptomics,  where Transformers are commonly implemented and other architectures are overlooked. Experimental results demonstrate that GNNs can achieve competitive performance compared with Transformers, with less consumption. This finding suggests a promising direction for designing new model structures in single-cell transcriptomics.
\end{enumerate}

\section{Related Works}

\noindent\textbf{Graph Neural Networks (GNNs).} 
GNNs~\cite{gori2005new,wu2020surveyGNN} were initially designed to process graph-structured data, focusing on capturing the topological interactions between nodes. Recently, numerous integrations were developed to enhance their functionality, including the combination with Transformers. Among these, 
Graph Convolutional Networks~\cite{henaff2015deep} introduced convolutions into GNNs, Graph Autoencoders~\cite{kipf2016variational} introduced the encoder-decoder framework to node learning,
Spatio-Temporal GNNs~\cite{yu2017spatio} introduced temporal modules to address the dynamic graphs, 
Graph Attention Networks~\cite{velivckovic2017GAT} incorporated attention mechanisms to allow for flexible feature aggregation from neighboring nodes, and Graph Transformers~\cite{yang2021graphformers} integrated Transformers to capture long-range dependencies. 
Although researchers have attempted to expand GNN applications with other structures and even combine Transformers, the intrinsic research between GNNs and Transformers, such as the consistency and distinctions between them, and whether GNNs can substitute Transformers under certain conditions, has not been fully explored.

Due to the multi-head self-attention mechanism, which requires computation across all token pairs in the input sequence, and large embedding layers, Transformers usually require higher computational resources and memory consumption, especially for long sequences or high-dimensional data~\cite{tay2022efficient,ansar2024survey}. GNNs typically perform computations on sparse graph structures, resulting in lower GPU consumption, especially when graphs have fewer nodes and edges~\cite{wu2020surveyGNN}. GNNs are often more memory-efficient than Transformers because they use local node connections without requiring attention calculations across all nodes. Moreover, GNNs avoid the need for attention map calculations, and further optimizations on the graph structure can significantly reduce computational complexity (FLOPs).

\smallskip
\noindent\textbf{Positions for Transformer.}
Positions are critical for Transformers to perform such functions, which are encoded as absolute position encoding in vanilla Transformer~\cite{vaswani2017attention}, GPT3~\cite{brown2020languageGPT3} and OPT~\cite{zhang2022opt}, or relative position encoding in T5~\cite{raffel2020exploringT5}, Rotary~\cite{su2024roformer} and BLOOM~\cite{le2023bloom}, or even no position encoding~\cite{haviv-etal-2022-transformer} which can implicitly encode the positions. Although the importance of position information for Transformer has been widely studied in various areas~\cite{dufter2022position,mao2022towards,junfengpositional}, the scenarios where positions are inherently absent have been rarely investigated. In this study, we do not discuss specific position encoding methods; instead, we examine the performance of Transformers when positional encoding information is entirely absent.

\smallskip
\noindent\textbf{Single-Cell Transcriptomics.} Single-cell transcriptomics, also known as single-cell RNA sequencing (scRNA-seq), was first introduced by Surani Lab~\cite{tang2009mrna}. It has since become a widely adopted tool in human health research, particularly for characterizing cell types across various organs~\cite{voigt2021single, ramachandran2020single}, exploring transcriptomic heterogeneity within similarly classified cells at different stages~\cite{kravets2020transcriptome, wheeler2020mafg} and elucidating temporal processes, such as human tissue development~\cite{olaniru2023single, collin2021single}. Due to the sequential nature of single-cell transcriptomic data, with each gene treated as a token, Transformers have been leveraged to address challenges in this field. The introduction of scBERT~\cite{scBERT} marked a pioneering effort to apply Transformer-based pre-training, utilizing masked token prediction. Subsequent studies have focused on expanding the dataset size~\cite{scGPT}, enhancing dataset diversity~\cite{genecompass2023}, and modifying Transformer architectures~\cite{scFoundation, geneFormer2023}. However, the potential of other architectures, such as Graph Neural Networks (GNNs), has yet to be fully explored in single-cell transcriptomics. In this paper, we propose the application of GNNs as a substitute for Transformers in this field, paying the way for future advancements in model development in single-cell transcriptomics.

\section{Method}
\subsection{Preliminaries}
\noindent\textbf{Sequential Dataset.} Considering a sequential dataset \(\mathcal{D} = \{\bm{x}_i\}_{i=1}^N\), where \(\bm{x}_i = (x_{i,1}, x_{i,2}, \ldots, x_{i,l})\) consists of $l$ tokens (nodes) from a predefined dictionary. The common self-supervised objective is \textit{Masked Token Prediction (MTP)}, also known as masked language modeling~\cite{devlin2018bert}. The input is masked \(\bm{x}_i\), where some tokens of \(\bm{x}_i\) are randomly masked, e.g., \(\tilde{\bm{x}}_i = (x_{i,1}, [mask], \ldots, x_{i,l})\). The output is the predictions for the masked tokens. To optimize a model with parameters $\theta$, the loss function for \textit{MTP} can be written as a classification loss for all masked tokens: 
\begin{align}
\label{eq:base_objective}
\begin{split}
\mathcal{L}_{\text{\textit{MTP}}}(\mathcal{D}, \theta) = - \frac{1}{N} &\sum_{i=1}^N \sum_{k} \mathbf{y}_{i,k}\log {p(\tilde{\bm{x}}_i, \theta)_k},
\end{split}
\end{align}
where \(\mathbf{y}_{i,k}\) is the one-hot token label for \(k\)-th masked token in \(\tilde{\bm{x}}_i\), and \(p(\tilde{\bm{x}}_i, \theta)_k\) is the predicted probability for \(k\)-th masked token in \(\tilde{\bm{x}}_i\).

For the calculation of \(p(\tilde{\bm{x}}_i, \theta)_k\), the input \(\tilde{\bm{x}}_i\) is initially processed by an embedding layer to generate embeddings $\bm{e}$, and then fed to feature extractors, such as GNNs and Transformers, to derive features $\bm{e}'$, and finally to calculate the classification loss. The whole process is written as follows:
\begin{gather}
\label{eq:softmax_calc1}
\bm{e} = (e_1, e_2, \ldots, e_l) = \mathbf{E_\theta}(\tilde{\bm{x}}_i), \\
\label{eq:softmax_calc2}
\bm{e}' = (e'_1, e'_2, \ldots, e'_l) = \mathbf{M_\theta}(\bm{e}), \\
\label{eq:softmax_calc3}
p(\tilde{\bm{x}}_i, \theta)_k = \frac{\exp(e'_{k} \cdot \bm{w})}{\sum_j \exp(e'_{k} \cdot \bm{w}_j)},
\end{gather}
where $\mathbf{E_\theta}$ denotes the embedding layer, note that the absolute position encoding is omitted here, $\mathbf{M_\theta}$ denotes GNN or Transformer, $\exp$ denotes the exponential function, \(\bm{w}\) is the classifier weight for token prediction, \(\cdot\) denotes dot product, and \(j\) indexes all tokens in the token dictionary.

\begin{figure*}[t]
\centering\includegraphics[width=\linewidth]{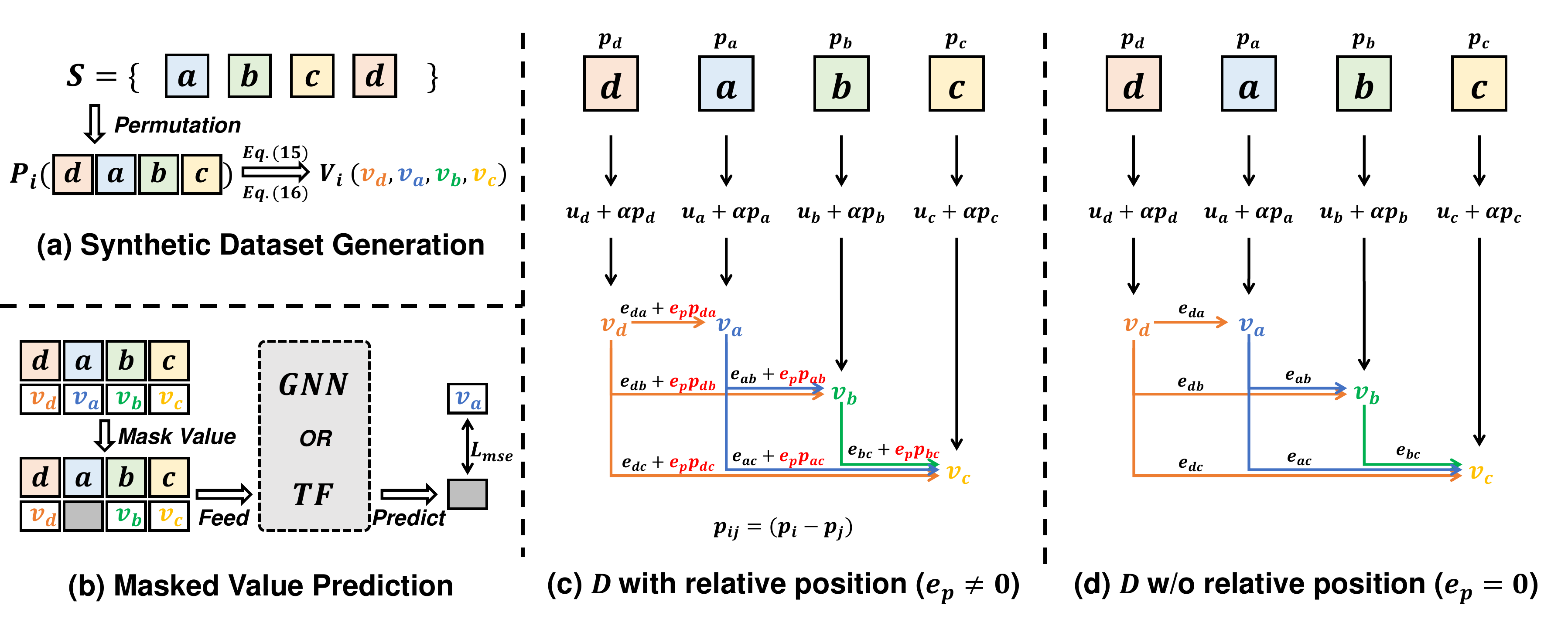}
    \caption{Illustrations for synthetic example. (a) outlines synthetic dataset generation. (b) shows the model training process. (c) and (d) illustrate generations for $\mathcal{D}$ by Eq.~\eqref{eq:gen_func1} with different handling of relative positions. Coefficient $e_p$ and Eq.~\eqref{eq:gen_func2} transform the obscure concept of relative positions into quantifiable values. Specifically, in (c), coefficients highlighted in red illustrate the impact of relative positions $p_{ij}$ on value generation, while such influence is eliminated by setting $e_p=0$ in (d).}
    \vspace{6pt}
    \label{fig:toy_gen}
\end{figure*}

\noindent\textbf{Sequence-like Dataset with Values.} Compared to sequential datasets, many sequence-like datasets, such as single-cell transcriptomics, do not have relative order but include additional values corresponding to each token: $\mathcal{D}\!=\!\{(\bm{x}_i, \bm{v}_i)\}_{i=1}^N$, where $(\bm{x}_i, \bm{v}_i) = (x_{i,1}, x_{i,2}, \ldots, x_{i,l}; v_{i,1}, v_{i,2}, \ldots, v_{i,l})$. Note that each token in $\bm{x}_i$ appears at most once and its position is casual, making traditional \textit{MTP} training infeasible. Instead, a variant known as \textit{Masked Value Prediction (MVP)}~\cite{scBERT}, as shown in Figure~\ref{fig:toy_gen}(b), is implemented for token learning, where values are masked to yield the input $(\bm{x}_i, \tilde{\bm{v}}_i) = (x_{i,1}, x_{i,2}, \ldots, x_{i,l}; v_{i,1}, [mask], \ldots, v_{i,l})$, and the output is the predictions for the masked values. The loss function of \textit{MVP} can be written as a regression for all masked values: 
\begin{align}
\label{eq:mvp_objective}
\mathcal{L}_{\text{\textit{MVP}}}(\mathcal{D}, \theta) = \frac{1}{N} \sum_{i=1}^N \sum_{k} \left\|v_{i,k} - f((\bm{x}_i, \tilde{\bm{v}}_i), \theta)_k\right\|^2,
\end{align}
where $v_{i,k}$ is the $k$-th masked value in $\tilde{\bm{v}}_i$ and $f((\bm{x}_i, \tilde{\bm{v}}_i), \theta)_k$ represents the corresponding predicted value.

For the calculation of $f((\bm{x}_i, \tilde{\bm{v}}_i), \theta)_k$, different from Eq.~\eqref{eq:softmax_calc1}, the embedding for $(\bm{x}_i, \tilde{\bm{v}}_i)$ is the aggregation (\eg, summation) of token and value embeddings. The whole process can be written as follows:
\begin{gather}
\label{eq:embed_process}
\bm{e} = \mathbf{E_\theta}(\bm{x}_i) + \mathbf{L_{1,\theta}}(\tilde{\bm{v}}_i), \\
\label{eq:feature_extraction}
\bm{e}' = (e'_1, e'_2, \ldots, e'_l) = \mathbf{M_\theta}(\bm{e}), \\
\label{eq:value_prediction}
f((\bm{x}_i, \tilde{\bm{v}}_i), \theta)_k = \mathbf{L_{2,\theta}}(e'_k),
\end{gather}
where $\mathbf{L_{1,\theta}}$ is a linear layer to encode the values into embeddings, $\mathbf{L_{2,\theta}}$ is a linear layer to predict the masked values from the extracted features, $d$ is the hidden dimension, $\mathbf{E_\theta}$ and $\mathbf{M_\theta}$ serves as the same roles described in Eq.~\eqref{eq:softmax_calc1} and Eq.~\eqref{eq:softmax_calc2}, respectively.

To explore the equivalence between GNNs and Transformers without relative positions, we adopt the setting {Sequence-like Dataset with Values}. Considering that the fundamental difference between these two architectures is encapsulated in Eq.~\eqref{eq:feature_extraction}, we will simplify the comparison by focusing on the different single-layer processes, excluding the common linear layers.

\medskip
\noindent\textbf{Transformers.} Transformers consist of a stack of attention layers, where the attention is dynamically computed for embeddings $\bm{e}$. Following Eq.~\eqref{eq:feature_extraction}, the attention mechanism can be written as:
\begin{align}
\bm{e}' &= \text{\text{Softmax}}\left((\bm{e}\mathbf{w}_q )(\bm{e}\mathbf{w}_k )^T\right) (\bm{e}\mathbf{w}_v ),
\end{align}
where $\mathbf{w}_q, \mathbf{w}_k, \mathbf{w}_v \in \mathbb{R}^{d \times d}$  are the linear layers for deriving the queries, keys, and values from $\bm{e} \in \mathbb{R}^{l \times d}$, respectively, $d$ is the hidden dimension, $l$ is the length of the sequence, and the normalization factor is omitted here for simplicity.

For each token embedding $e_i$, its attention weights $\bm{a_{i}}$ for other tokens in the sequence can be written as:
\begin{align}
\bm{a_{i}} &= \frac{(\exp(s_{i1}),\exp(s_{i2}),\ldots,\exp(s_{il}))}{\sum_{k=1}^l \exp(s_{ik})},
\label{att_tf}
\end{align}
where $s_{ij} = (e_i\mathbf{w}_q ) (e_j\mathbf{w}_k )^T$, $l$ is the length of the sequence, and $\exp$ denotes the exponential function.

According to Eq.~\eqref{att_tf}, in Transformers, the attention between two tokens, \emph{i.e.}, the information transfer between them, is computed dynamically based on their current embeddings, which ensures the relative positions can be encoded into attentions. 

\medskip
\noindent\textbf{Graph Neural Networks (GNNs).} GNNs use graph topology to determine the information transfer, typically utilizing an adjacency matrix. By treating tokens in the sequence as nodes in an underlying graph, Eq.~\eqref{eq:feature_extraction} can be written as:
\begin{align}
\bm{e}' &= \text{Softmax}(\mathbf{A}) \bm{e},
\end{align}
where $\mathbf{A} \in \mathbb{R}^{l \times l}$ denotes the adjacency matrix for the graph of input tokens, 
and $\text{Softmax}$ is applied for normalization.

For each token embedding $e_i$, its interactions $\bm{a_{i}}$ with other tokens can be written as:
\begin{align}
\bm{a_{i}} &= \frac{(\exp(\mathbf{A}_{i1}),\exp(\mathbf{A}_{i2}),\ldots,\exp(\mathbf{A}_{il}))}{\sum_{k=1}^l \exp(\mathbf{A}_{ik})},
\label{att_gnn}
\end{align}
where $\mathbf{A}_{ij}$ represents the edge from node $i$ to node $j$ in the graph, and $l$ is the length of the sequence.

According to Eq.~\eqref{att_gnn}, in GNNs, the information transfer between two nodes is determined by $\mathbf{A}$, which is invariant across different samples. Therefore, GNNs are unable to capture the changes in relative positions between the same node pairs. 

\medskip
\noindent\textbf{Comparisons of Empirical Risks.} To compare the empirical risks, $\mathcal{R}$, which can be calculated by Eq.~\eqref{eq:mvp_objective}, we assume that for sample pair \((\bm{x},\bm{v})\), the $k$-th value is defined as the summation of the influence of token, other values and other values considering relative positions:
\begin{align}
\label{value_gen_func}
 v_k = h_1(x_k) + {\textstyle\sum_j} h_2(v_j) + {\textstyle\sum_j} h_3(p_{kj}v_j),
\end{align}
where $h_1$, $h_2$ and $h_3$ are inherent functions for value generation, $j$ indexes all tokens except $k$ in the sample, $p_{kj}$ is the relative positions between $k$ and $j$.  

Assuming that Transformers and GNNs have a similar ability to learn the first two terms and $\mathcal{R}$ for $h_3$ across all $k$-th values can be written as:
\begin{align}
 \mathcal{R}_{h_3} &= {\textstyle\sum_{\mathcal{D}}\sum_j} \|h_3(p_{kj}v_j) - f(p_{kj},v_j)\|^2,
\end{align}
where $f$ is the function with learnable parameters, which can be written as $f(s_{kj}v_j)$ in Transformer and $f(\mathbf{A}_{kj}v_j)$ in GNN. Note that $s_{kj} = (e_k\mathbf{w}_q ) (e_j\mathbf{w}_k )^T$ is the function of $p_k$ and $p_j$, while $\mathbf{A}_{kj}$ does not, indicating that Transformers could achieve lower $\mathcal{R}_{h_3}$ with varying $p_{kj}$. When eliminating the relative positions $p_{kj}$, the function $h_3$ will degenerate to $h_2$ and thus can be minimized by both of the two models. More details are in the Appendix.

Although we have elucidated the primary differences between Transformers and GNNs, the theoretical justification for their equivalence remains challenging due to variations in model structures and optimizations. Here, we adopt a more practical and concise approach by designing a synthetic example explicitly involving relative positions in data generation to validate their equivalence.

\newcommand{\INPUT}{\item[\textbf{Input:}]}
\newcommand{\OUTPUT}{\item[\textbf{Output:}]}
\newcommand{\STAGEONE}{\item[\textbf{Stage 1:}]}
\newcommand{\STAGETWO}{\item[\textbf{Stage 2:}]}
\begin{algorithm}[t]
    \caption{Synthetic Example}
    \begin{algorithmic}

    \setlength\itemsep{0.2em} 
    \STAGEONE Synthetic Datasets Generation
    \INPUT Token set \(S\!=\!\{a, b, c, d\}\), positional coefficient $e_p\in\{0, 0.2,\ldots\}$
    \OUTPUT Datasets \(\{\mathcal{D}_{k}\}\) corresponding to $e_p=k$
    \FOR{ $e_p~\textbf{in}~\{0, 0.2,\ldots\}$}
        \STATE Initialize dataset \(\mathcal{D}_{k}\leftarrow\varnothing\) 
        \WHILE{$|\mathcal{D}_k|<N$}
        \STATE Sample permutation $P_i$ from \( \text{Perm}(S) \)
        \STATE Generate values $V_i$ by Eq.~\eqref{eq:gen_func1} and Eq.~\eqref{eq:gen_func2}
        \STATE \(\mathcal{D}_k \leftarrow \mathcal{D}_k \cup \{(P_i, V_i)\}\)
        \ENDWHILE
    \ENDFOR
    \STAGETWO Verification for GNNs and Transformers
    \INPUT Datasets \(\{\mathcal{D}_{k}\}\), model parameters $\theta_{gnn}$ and $\theta_{tf}$
    \OUTPUT Testing losses \(\mathcal{L}_{gnn,k},\mathcal{L}_{tf,k}\) for each \(\mathcal{D}_{k}\)
    \FOR{\(\mathcal{D}_{k}\) \textbf{in}  \(\{\mathcal{D}_{k}\}\) }
        \STATE Split \(\mathcal{D}_k\) into \(\mathcal{D}_{tr,k}\) and \(\mathcal{D}_{te,k}\)
        \STATE Randomly initialize $\theta_{gnn}$ and $\theta_{tf}$
        \STATE Train $\theta_{gnn}$ and $\theta_{tf}$  on $\mathcal{D}_{tr,k}$ by Eq.~\eqref{eq:mvp_objective}, respectively
        \STATE Test $\theta_{gnn}$ and $\theta_{tf}$ on $\mathcal{D}_{te,k}$ to derive \(\mathcal{L}_{gnn,k},\mathcal{L}_{tf,k}\)
    \ENDFOR
    \end{algorithmic}
    \label{algo:toy example}
\end{algorithm}

\subsection{Synthetic Example}
\label{toy_example}
We design and generate a synthetic dataset to further evaluate the equivalence between Graph Neural Networks and Transformers with or without relative positions. As illustrated in Figure~\ref{fig:toy_gen}(a), we consider the classic Masked Value Prediction task for sequence-like data.

First, we define a set \(S\) containing four tokens \(\{a, b, c, d\}\), and use them to generate various permutations $P$ with distinct elements. Then, we generate corresponding quadruple values $V$ based on the sequence and the defined formula, resulting in data pairs, such as \(x_i = (P_i, V_i) = (({d, a, b, c}), (1.0, 1.5, 3.0, 2.5))\). As illustrated in Figure~\ref{fig:toy_gen}(c) and Figure~\ref{fig:toy_gen}(d), we can set the importance of relative position information in the dataset by controlling the coefficient \(e_p\), thus enabling us to validate the position encoding capabilities of GNNs and Transformers. Finally, using the same settings and the same parameters scale, we train both networks and thus can determine their equivalence by comparing the test losses.
Here, inspired by  Eq.~\eqref{value_gen_func}, we use the sampled permutation \(({d, a, b, c})\) to generate the corresponding values \(({v_d, v_a, v_b, v_c})\) as an example to introduce our predefined functions:
\begin{equation}
\begin{aligned}
    v_d &= u_d + \alpha p_d, \\
    v_a &= u_a + \alpha p_a + \beta(d,a,p_d,p_a) \cdot v_d, \\
    v_b &= u_b + \alpha p_b + \textstyle \sum_{i \in \{d, a\}} \beta(i,b,p_i,p_b) \cdot v_{i}, \\
    v_c &= u_c + \alpha p_c + \textstyle \sum_{i \in \{d, a, b\}} \beta(i,c,p_i,p_c) \cdot v_{i},
    \label{eq:gen_func1}
\end{aligned}
\end{equation}
where $p_i$ is the absolute position of $i$, \eg, $p_a\!=\!2$ in the permutation \(({d, a, b, c})\), $\alpha$ is the coefficient for the absolute position, $u_i\sim \mathcal{N}(\mu_i, \sigma_i)$ is the exogenous value of $i$, sampled from a Gaussian distribution corresponding to the token $i \in \{a,b,c,d\}$, $\beta$ is the function to encode the relative positional information, which is written as: 
\begin{equation}
    \beta(i,j,p_i,p_j) = e_{ij} + e_p \cdot (p_i - p_j),
    \label{eq:gen_func2}
\end{equation}
where $e_{ij}$ is the static coefficient for the information transfer from the token $i$ to $j$, $e_p$ is the coefficient to determine the influence of relative position in this dataset, \ie, there is no relative position influence when $e_p\!=\!0$.


\section{Experiments}
\subsection{Synthetic Experiments}

\noindent\textbf{Dataset.} We followed the procedure in Algorithm~\ref{algo:toy example} to initially permute the set $\mathcal{S}$ to derive $\bm{x}$, and subsequently used Eq.~\eqref{eq:gen_func1} and Eq.~\eqref{eq:gen_func2} to generate values for each element of $\bm{x}$. Specifically, the exogenous values $\{\mu_a, \mu_b, \mu_c, \mu_d\}$ were set to $\{-0.5, -0.25, 0.25, 0.5\}$, respectively, and the absolute
positions $\{p_a, p_b, p_c, p_d\}$ were assigned the values $\{1, 2, 3, 4\}$, respectively. We set $\alpha$ to $-0.1$ and uniformly sampled $\{e_{ij},i\in\{a,b,c,d\},j\in\{a,b,c,d\}\}$ from the interval $(0.2,0.5)$, respectively, in each experimental setting. The parameter $e_p$, which controls the influence of relative positions, was restricted to the interval $(0,1)$, and $\sigma_i=\sigma$ for $i \in \{a, b, c, d\}$ was set according to the settings to control the variance of the data. For each experimental trial under a given setting of $(\sigma, e_p)$, new datasets were generated, consisting of 20,000 samples, which were split into a training set and a test set at ratios of 70\% and 30\%, respectively.

\smallskip
\noindent\textbf{Implementation Details.} To ensure fair comparisons, GNNs and Transformers were tested under identical experimental settings across datasets generated for each $(\sigma, e_p)$. These settings included masking one random value for loss calculation by Eq.~\eqref{eq:mvp_objective}, using batch size as 256, 200 training epochs, and Adam~\cite{kingma2014adam} optimizer with a learning rate as 0.001. The best test loss was recorded as the performance for each model. To minimize the influence of parameter scale, single-layer GNNs and Transformers were implemented with an equal hidden dimension of 8. $\mathbf{A}$ was implemented as a trainable matrix. Input and output layers were implemented as single linear layers where $d=8$.


\begin{table}[t]
\centering
\caption{Comparisons of test losses in synthetic example between Transformers and GNNs with varying $e_p$ and $\sigma$. $r_\Delta\!=\! \mathcal{L}_{GNN}/\mathcal{L}_{TF}$ indicates the relative loss differences between them. Avg denotes the averaged result across $\sigma$. For $r_\Delta$, a larger number denotes Transformer is better, and for others, a smaller number is better.}
    \renewcommand{\arraystretch}{1.15}
     \scalebox{1.0}{
\begin{tabular}{c|c|ccccc|c}

\toprule
\(e_p \backslash \sigma\)  &  & 0.1 & 0.2 & 0.3 & 0.4 & 0.5 & Avg \\ 
 \hline
\multirow{3}{*}{0} & TF &  0.009 & 0.034 & 0.077& 0.134 & 0.207 & - \\
&GNN         & 0.013 & 0.040 & 0.085 & 0.150 & 0.225 & -\\
& $r_\Delta$  & 1.444 & 1.176 & 1.104 & 1.119 & 1.087 & 1.186 \\
\hline
\multirow{3}{*}{0.1} & TF & 0.010 & 0.030 & 0.069 & 0.119 & 0.217 & - \\
&GNN         & 0.026 & 0.057 & 0.098 & 0.159 & 0.242 & -\\
& $r_\Delta$  & 2.600 & 1.900 & 1.420 & 1.336 & 1.115 & 1.674 \\
\hline
\multirow{3}{*}{0.2} & TF & 0.009 & 0.035 & 0.063 & 0.133 & 0.168& -\\
&GNN         & 0.031 & 0.103 & 0.122 & 0.203 & 0.299 & -\\
& $r_\Delta$  & 3.444 & 2.943 & 1.937 & 1.526 & 1.780 & 2.326\\
\hline
\multirow{3}{*}{0.3} & TF & 0.009 & 0.028 & 0.059 & 0.135 & 0.197 & -  \\
&GNN         & 0.055 & 0.104 & 0.174 & 0.278 & 0.399 & -\\
& $r_\Delta$  & 6.111 & 3.714 & 2.949 & 2.059 & 2.025 & 3.372 \\
\hline
\multirow{3}{*}{0.4} & TF & 0.014 & 0.028 & 0.059 & 0.133 & 0.156 & -\\
&GNN         & 0.096 & 0.166 & 0.269 & 0.406 & 0.584 & -\\
& $r_\Delta$  & 6.857 & 5.929 & 4.559 & 3.053 & 3.744 & 4.828\\
\hline
\multirow{3}{*}{0.5} & TF & 0.010& 0.027 & 0.085 & 0.163& 0.193 & - \\
&GNN         &0.163 & 0.267 & 0.421  &0.621 & 0.884 & - \\
& $r_\Delta$  & 16.30 &9.889 & 4.953& 3.810 & 4.580 & 7.906\\
\bottomrule
\end{tabular}
}

\label{toy_table}
\end{table}
\smallskip
\noindent\textbf{Results and Analysis.}

\begin{figure}[t!]
\centering\includegraphics[width=\linewidth]{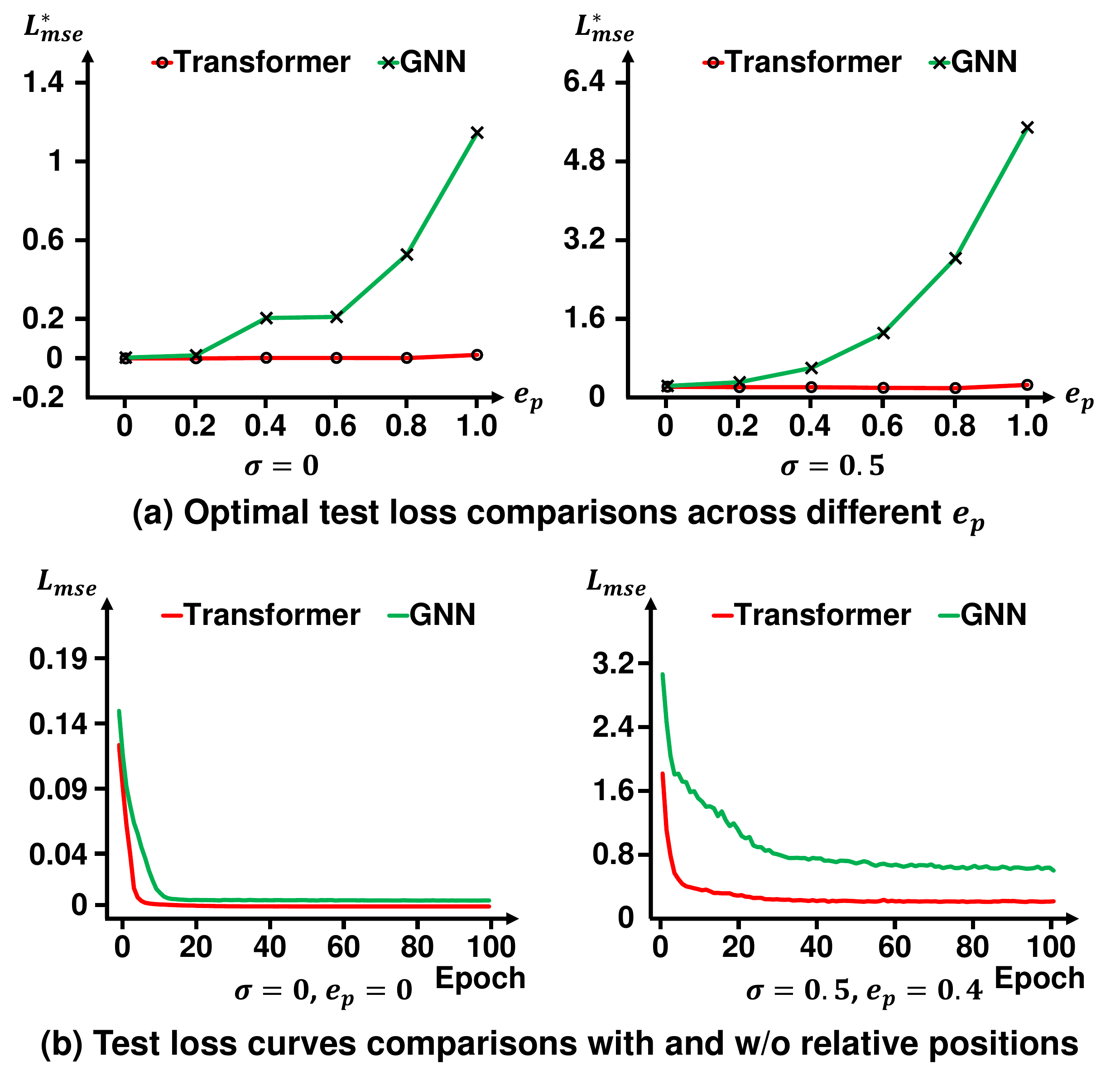}
    \caption{Visualizations of test losses for Transformers (red) and GNNs (green) in synthetic example. (a) Comparisons for optimum test losses $L_{mse}^*$ across different $e_p$. $e_p=0$ indicates no influence of relative positions while increasing $e_p$ denotes greater influence. $\sigma$ represents data variance. (b) Comparisons for test loss curves. Left: without relative positions. Right: with relative positions.}
    \label{fig:loss_toy}
\end{figure}
\noindent\textit{Q1.} \textit{Does the superiority of Transformers over GNNs originate from their capacity to capture relative positions?}

As shown in the last column of Table~\ref{toy_table}, with the increase of $e_p$, the indicator of influence for relative positions, Transformers increasingly outperform GNNs, where averaged $r_\Delta$ increased from $1.186$ to $9.670$ (larger value denotes Transformer is better). This affirms the hypothesis in \textit{Q1}. Specifically, as $e_p$ increases, the optimal test losses for Transformers remain consistent across each $\sigma$, indicating their optimization is robust for relative positions. In contrast, the optimal test losses for GNNs worsen with rising $e_p$ values, indicating their lack of capability in handling relative positions. Additionally, Figure~\ref{fig:loss_toy}(a) clearly illustrates this trend, showing that as $e_p$ ascends, Transformers (red) maintain a stable horizontal line, whereas GNNs (green) escalate, increasingly expanding the performance gap. The right of Figure~\ref{fig:loss_toy}(b) also highlights the optimization differences between the two models in the presence of relative positions.

\begin{table*}[t!]
    \centering
        \caption{Test Accuracy (\%) of Transformers and GNNs on five transcriptomic classification datasets. Numbers in the Settings column denote the hidden dimension of the models and the sequence length of inputs, respectively. Mem denotes the occupied GPU memory by a single sample in the training. Results are averaged over five independent trials (mean $\pm$ std). 
    }
    \renewcommand{\arraystretch}{1.16}
    \scalebox{1.1}{
      \begin{tabular}{c|c|cc|ccccc|c}
        \toprule
        Settings & Model & FLOPs & Mem & Lept & Sclerosis & Pancreas & Lupus & Dengue & Average \\
        \hline\hline
        \multirow{2}{*}{256/512} & TF & 2.01G & 169.96M &  95.17$\pm$0.33 & 89.18$\pm$0.34 & 97.88$\pm$0.30 & 77.80$\pm$0.17 & 86.36$\pm$0.20 & 89.28 \\
        & \cellcolor{gray!15}GNN & \cellcolor{gray!15}0.76G & \cellcolor{gray!15}27.50M & \cellcolor{gray!15}{94.83$\pm$0.12} & \cellcolor{gray!15}{89.17$\pm$0.21} & \cellcolor{gray!15}{97.05$\pm$0.16} & \cellcolor{gray!15}{77.72$\pm$0.20} & \cellcolor{gray!15}{86.71$\pm$0.18} & \cellcolor{gray!15}{89.10} \scriptsize{\textcolor{black}{-0.14}} \\
        \hline\hline
        \multirow{2}{*}{128/512}   & TF & 0.70G & 147.52M & 94.55$\pm$0.18 & 87.64$\pm$0.23 & 95.92$\pm$0.38 & 72.09$\pm$0.35 & 84.88$\pm$0.21 & 87.00 \\
        & \cellcolor{gray!15}GNN & \cellcolor{gray!15}0.41G & \cellcolor{gray!15}19.41M & \cellcolor{gray!15}{93.13$\pm$0.20} & \cellcolor{gray!15}{86.34$\pm$0.26} & \cellcolor{gray!15}{94.23$\pm$0.15} & \cellcolor{gray!15}{73.23$\pm$0.30} & \cellcolor{gray!15}{83.76$\pm$0.16} & \cellcolor{gray!15}{86.14} \scriptsize{\textcolor{black}{-0.86}} \\
         \hline
        \multirow{2}{*}{512/512}   & TF & 6.44G & 216.04M & 95.65$\pm$0.14 & 89.10$\pm$0.16 & 98.29$\pm$0.09 & 79.51$\pm$0.21 & 87.28$\pm$0.15 & 89.96 \\
        & \cellcolor{gray!15}GNN & \cellcolor{gray!15}1.77G & \cellcolor{gray!15}44.34M & \cellcolor{gray!15}{95.33$\pm$0.17} & \cellcolor{gray!15}{90.07$\pm$0.41} & \cellcolor{gray!15}{97.82$\pm$0.13} & \cellcolor{gray!15}{78.98$\pm$0.36} & \cellcolor{gray!15}{87.47$\pm$0.29} & \cellcolor{gray!15}{89.93} \scriptsize{\textcolor{black}{-0.03}} \\
        \hline\hline
        \multirow{2}{*}{256/256}   & TF & 0.81G & 52.29M & 94.82$\pm$0.22 & 88.57$\pm$0.25 & 98.02$\pm$0.31 & 76.42$\pm$0.26 & 85.60$\pm$0.09 & 88.69  \\
        & \cellcolor{gray!15}GNN & \cellcolor{gray!15}0.24G & \cellcolor{gray!15}9.41M & \cellcolor{gray!15}{94.60$\pm$0.17} & \cellcolor{gray!15}{88.10$\pm$0.26} & \cellcolor{gray!15}{96.93$\pm$0.19} & \cellcolor{gray!15}{76.48$\pm$0.41} & \cellcolor{gray!15}{86.11$\pm$0.28} & \cellcolor{gray!15}{88.44} \scriptsize{\textcolor{black}{-0.25}}\\
        \hline
        \multirow{2}{*}{256/1024}  & TF & 5.64G & 588.14M & 95.52$\pm$0.24 & 88.48$\pm$0.32 & 98.12$\pm$0.18 & 78.19$\pm$0.41 & 86.69$\pm$0.21 & 89.40  \\
        & \cellcolor{gray!15}GNN & \cellcolor{gray!15}2.63G & \cellcolor{gray!15}76.63M & \cellcolor{gray!15}{94.89$\pm$0.13} & \cellcolor{gray!15}{89.39$\pm$0.20} & \cellcolor{gray!15}{97.23$\pm$0.17} & \cellcolor{gray!15}{78.61$\pm$0.18} & \cellcolor{gray!15}{85.78$\pm$0.39} & \cellcolor{gray!15}{89.18} \scriptsize{\textcolor{black}{-0.22}}\\
        \hline\hline
        \multirow{2}{*}{Average} & TF & - & - & 95.14& 88.59 & 97.64 & 76.80 & 86.16 & 88.87\\
        & \cellcolor{gray!15}GNN & \cellcolor{gray!15}- & \cellcolor{gray!15}- & \cellcolor{gray!15}94.56 \scriptsize{\textcolor{black}{-0.58}} & \cellcolor{gray!15}88.61 \scriptsize{\textcolor{black}{+0.02}} & \cellcolor{gray!15}96.65 \scriptsize{\textcolor{black}{-0.99}} & \cellcolor{gray!15}77.00 \scriptsize{\textcolor{black}{+0.20}} & \cellcolor{gray!15}85.96 \scriptsize{\textcolor{black}{-0.20}} & \cellcolor{gray!15}88.56 \scriptsize{\textcolor{black}{-0.31}}\\
        \bottomrule
      \end{tabular}
    }

    \label{tab:main_results}
    
\end{table*}


\noindent\textit{Q2.} \textit{Do Transformers and GNNs demonstrate equivalence when relative positions are absent?} 

As shown in Table~\ref{toy_table} and Figure~\ref{fig:loss_toy}(a), when relative positions are absent, \ie, $e_p=0$, the performance gap between Transformers and GNNs is negligible, with the $r_\Delta$ consistently close to 1 and the difference reaching as low as $8.7$\%. This indicates that Transformers and GNNs can be regarded as equivalent. The test loss curves presented in Figure~\ref{fig:loss_toy}(b) Left further confirm that both models ultimately converge to comparable optimal losses.

\subsection{Experiments for Transcriptomics}
We selected single-cell transcriptomics that aligns with the sequence-like dataset with values settings, where there are no positions, to conduct the large-scale experiments. We followed the procedure~\cite{scGPT} to pre-train models and then assessed them by downstream classification tasks to demonstrate the equivalence between Transformers and GNNs. Note that more details about the dataset and implementations for single-cell transcriptomics are in the Appendix.

\begin{figure}[b]
\centering\includegraphics[width=\linewidth]{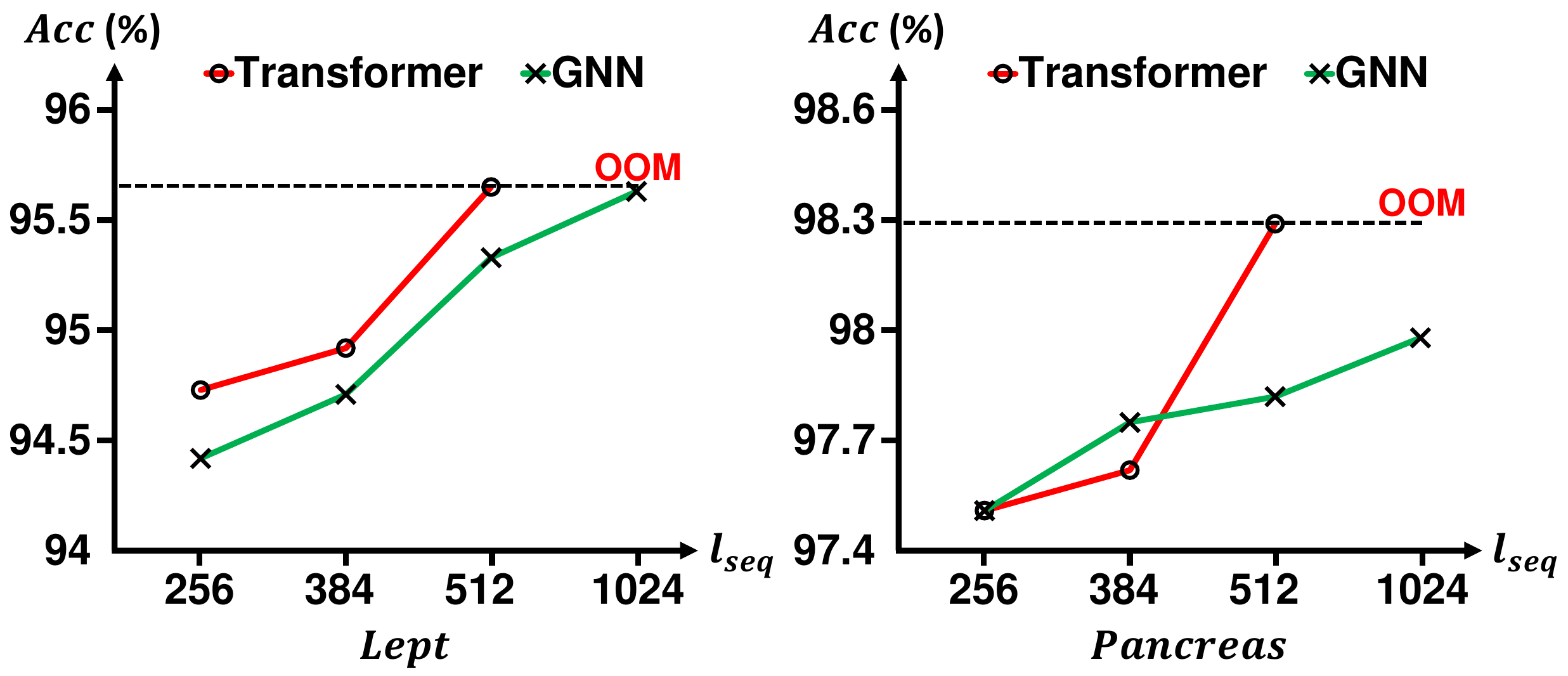}
    \caption{Accuracy (\%) comparisons of Transformers (red) and GNNs (green) with varying sequence length $l_{seq}$ in two transcriptomic datasets where Transformers show the largest performance advantage over GNNs. OOM denotes out-of-memory errors due to large sequence length.}
    \label{fig:oom}
\end{figure}

\begin{figure*}[t]
\centering\includegraphics[width=\linewidth]{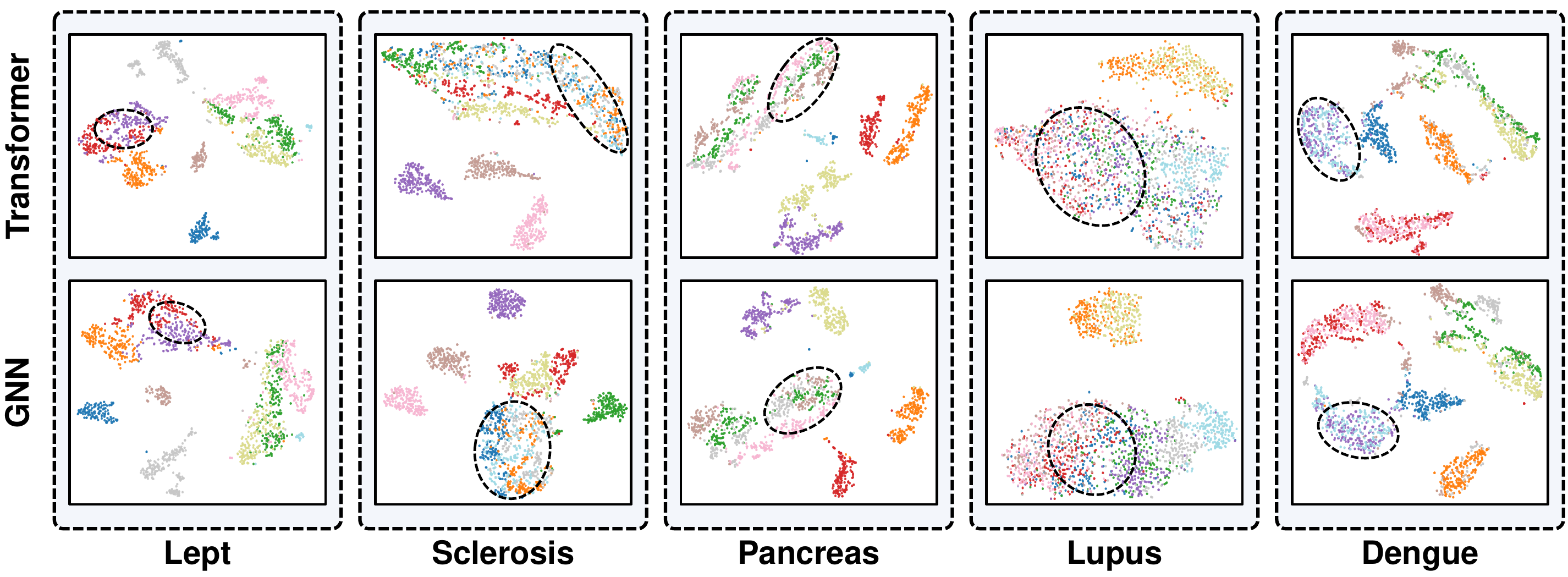}
    \caption{t-SNE visualization of features extracted by Transformers (Top) and GNNs (Bottom) across five transcriptomic datasets. For each dataset, only 10 classes are displayed for brevity, and black dashed circles highlight areas where both models exhibit similar confusion.}
    \label{fig:tsne}
\end{figure*}

\smallskip
\noindent\textbf{Dataset.} 

The transcriptomic dataset, drawn from the CELLxGENE collection~\cite{czi2023cz}, comprises human cells, featuring gene and expression pairs. We randomly sampled 1.8 million data points for pretraining. 

For downstream classification, we collected five transcriptomic datasets for comprehensive evaluations: Lept~\cite{remsik2023leptomeningeal} with 20,676 samples across 10 classes; Sclerosis~\cite{schirmer2019neuronalMS} with 21,312 samples and 18 classes; Pancreas~\cite{chen2023transformerPanc} with 14,818 samples and 14 classes; Lupus~\cite{perez2022single}, with 50,000 samples and 14 classes; and Dengue~\cite{ghita2023global} with 50,000 samples and 23 classes.

\smallskip
\noindent\textbf{Implementation Details.} 
To ensure fair comparisons, both GNNs and Transformers were configured under uniformly applied experimental settings, including a 6-layer model structure, a hidden dimension size of 256, a batch size of 256, an input gene length set as 512, and 10 training epochs. An Adam optimizer with a learning rate of 0.0002 was employed during pre-training. For GNNs, due to the lack of a pre-defined graph, the adjacency matrix $\mathbf{A}$ was modeled as a trainable matrix. Note that since the gene dictionary is extremely large, reaching up to $60,000$, optimizing such an adjacency matrix is clearly impractical. Therefore, we employed a commonly used matrix factorization method, optimizing two matrices of size $d \times s$, where $s$ represents the size of the gene dictionary and $d=48$ is the dimension for the reduced feature space, significantly reducing the number of parameters. The feature extractor was frozen for downstream classification tasks, and a single linear layer was trained as the classifier to fully evaluate pre-training performance. The downstream datasets were split into 70\% as the training set and 30\% as the testing set. We applied the settings, including 50 training epochs, and a batch size of 64, using the Adam optimizer with a learning rate of 0.005 for evaluating both of the two models. Each experiment was independently replicated five times using the same hardware, with the averaged results subsequently reported.


\smallskip
\noindent\textbf{Results and Analysis.} 

\noindent\textit{Q1.} \textit{Could GNNs substitute Transformers in single-cell transcriptomics? (\ie, whether they achieve equivalent performance?)}

The main results in the single-cell transcriptomics experiments are shown in Table~\ref{tab:main_results}. Under various training settings, GNNs and Transformers achieve similar classification accuracy, where the maximum gap is 0.86\% and the minimum is only 0.03\%, which is negligible. In the comparison across datasets, GNNs slightly outperform on Sclerosis and Lupus, and Transformers outperform on others, with only minor differences noted in both cases. In summary, the overall difference, \ie the classification accuracies averaged over all experiments, between these two models is only 0.31\%, effectively demonstrating their equivalence when training on single-cell transcriptomic datasets, \ie, where there are no positions, and of course, there are relative positions either.

\noindent\textit{Q2.} \textit{How do GNNs and Transformers perform in qualitative assessments?} 

As shown in Figure~\ref{fig:tsne}, GNNs and Transformers show high similarity in their extracted features. For example, in the Lept dataset, both models distinctly segregate the grey, brown, and blue classes in the t-SNE visualization, and display similar confusion between the red and purple classes (indicated by black dashed circles). This phenomenon is consistently observed across all datasets. Such qualitative results further demonstrate the equivalence between Transformers and GNNs in single-cell transcriptomic training without relative positions between tokens.


\noindent\textit{Q3.} \textit{Is it practical to employ GNNs as an alternative to Transformers?} 

Since their performance is quite similar, they can readily substitute for each other in datasets without positions. Considering computational resource consumption, as shown in Table~\ref{tab:main_results}, in our experiments on the transcriptomic dataset, under the same training settings, the FLOPs of Transformers range from 1.71 to 3.64 times those of GNNs, and the GPU memory usage during training is 4.87 to 7.68 times greater, establishing GNNs as the more computation-efficient choice. As shown in Figure~\ref{fig:oom}, GNNs can still improve performance by increasing the input length to chase Transformers, whereas Transformers may encounter out-of-memory (OOM) problems. These results suggest that GNNs are promising alternatives to Transformers in single-cell transcriptomics and other similar datasets where there are no positions.

\section{Conclusion}
Since the fundamental similarities between Transformers and vanilla Graph Neural Networks in node processing and feature interactions, we intuitively explored their similarities and differences and thus formulated their node interaction mechanisms. Then, we identified the ability to encode relative positions as their primary distinction, which was demonstrated by our designed synthetic experiments. Such differences prompted us to consider whether they remain equivalent in the absence of positions and whether we could replace Transformers with GNNs, especially in the fields where there is no relative positions between tokens, such as single-cell transcriptomics. After conducting extensive experiments with both synthetic datasets and large-scale position-agnostic datasets, \ie, single-cell transcriptomics, we validated that GNNs can indeed serve as an effective replacement for Transformers, achieving similar performance. This finding opens up new possibilities for researchers in the field of single-cell transcriptomics, suggesting that GNN-like models may offer a more efficient and innovative model architecture for advancing research, replacing the traditional reliance on Transformers in this domain. For our future work, we will build upon the current conclusions to explore the potential of advanced GNNs for single-cell transcriptomics. Additionally, we will investigate other similar datasets without positions to further demonstrate our conclusions between GNNs and Transformers.

\bibliographystyle{IEEEtran}
\bibliography{IEEEabrv,main}

\end{document}